\documentclass{article}
\usepackage{PRIMEarxiv}

\usepackage{hyperref}
\usepackage{color}
\usepackage{pifont}
\usepackage[many]{tcolorbox} 
\usepackage{paralist}
\usepackage{subfigure}
\usepackage{cleveref}
\usepackage{microtype}
\usepackage{fancyhdr}       
\usepackage{graphicx}       
\usepackage{xcolor}
\usepackage{multirow}

\title{
Identifying Uncertainty in Self-Adaptive Robotics with Large Language Models
}

\author{
  Hassan Sartaj \\
  Simula Research Laboratory \\
  Oslo, Norway\\
  \texttt{hassan@simula.no} \\
   \And
  Jalil Boudjadar \\
  Aarhus University \\
  Aarhus, Denmark\\
  \texttt{jalil@ece.au.dk} \\
  \And
  Mirgita Frasheri \\
  Aarhus University \\
  Aarhus, Denmark\\
  \texttt{mirgita.frasheri@ece.au.dk} \\
  \And
  Shaukat Ali \\
  Simula Research Laboratory \\
  Oslo, Norway\\
  \texttt{shaukat@simula.no} \\
  \And
  Peter Gorm Larsen \\
  Aarhus University \\
  Aarhus, Denmark\\
  \texttt{pgl@ece.au.dk} \\
}

\begin{document}
\maketitle

\begin{abstract}
Future self-adaptive robots are expected to operate in highly dynamic environments while effectively managing uncertainties. 
However, identifying the sources and impacts of uncertainties in such robotic systems and defining appropriate mitigation strategies is challenging due to the inherent complexity of self-adaptive robots and the lack of comprehensive knowledge about the various factors influencing uncertainty. 
Hence, practitioners often rely on intuition and past experiences from similar systems to address uncertainties. 
In this article, we evaluate the potential of large language models (LLMs) in enabling a systematic and automated approach to identify uncertainties in self-adaptive robotics throughout the software engineering lifecycle. 
For this evaluation, we analyzed 10 advanced LLMs with varying capabilities across four industrial-sized robotics case studies, gathering the practitioners’ perspectives on the LLM-generated responses related to uncertainties. 
Results showed that practitioners agreed with 63--88\% of the LLM responses and expressed strong interest in the practicality of LLMs for this purpose.
\end{abstract}


\section{Introduction}
Self-adaptive robotics refers to robotic systems that autonomously adjust their behavior, configuration, or decision-making processes in response to environmental changes and unforeseen circumstances~\cite{larsen2024robotic}. 
A fundamental framework for enabling self-adaptation in such systems is the MAPE-K (Monitoring, Analysis, Planning, Execution, and Knowledge) loop, which employs advanced techniques such as artificial intelligence and data analysis to allow robots to continuously collect and analyze data, plan and validate actions, execute adaptive behaviors, and refine decisions using a dynamic knowledge base~\cite{kephart2003vision}. 
At this level of autonomy, these robots must ensure dependability while effectively managing uncertainty and addressing ethical considerations to operate reliably in real-world environments.

A key challenge in self-adaptive robotics is managing uncertainty throughout the entire engineering lifecycle, from initial design to active operations~\cite{weyns2023towards}. 
Uncertainty may arise from various sources, such as unpredictable environmental conditions, sensor and actuator noise, and human-robot and robot-robot interactions.  
Such factors directly influence a robot’s dependability, ultimately impacting its overall performance and decision-making capabilities.  
A crucial step in managing uncertainty within these systems is identifying the sources and potential impacts of uncertainties at early stages of the robotic software engineering lifecycle~\cite{weyns2023towards}. 
However, it is challenging due to the inherent complexity of self-adaptive robots, including their complex interactions, evolving configurations, and adaptive behaviors, as well as the lack of comprehensive knowledge about unpredictable operational contexts and environmental dynamics. 
Consequently, practitioners often depend on their intuition and insights from previous experiences with similar systems to identify and manage uncertainties in self-adaptive robotics.

The growing adoption of large language models (LLMs) with advanced text capabilities, such as generation, analysis, and comprehension~\cite{LMMsDawn2023}, extends to various fields, including software engineering~\cite{hassan2024rethinking}.  
In this article, we explore LLMs' ability to assist practitioners in identifying uncertainties when developing self-adaptive robots throughout the entire software engineering lifecycle, aiming to provide a more systematic and comprehensive approach that improves over the current practice of relying primarily on intuition and experience. 

For evaluation, we selected 10 advanced LLMs with varying capabilities: Gemini Pro 1.5, Perplexity Sonar, Nemotron 70B, Nova Pro, Mistral Large 2411, LLama 3.3 70B, o1 Preview, GPT-4o, Gemini Flash 2.0, and Claude 3.5 Sonnet. 
We evaluated these LLMs across four real-world industrial robotic use cases: Industrial Disassembly Robot (IDR), Warehouse Robotic Swarm (WRS), Prolonged Hull of an Autonomous Vessel (PHAV), and Human-Robotic Interaction (HRI). 
To evaluate the LLM-generated responses regarding uncertainties, we conducted a study with practitioners from each case to collect their feedback. 
Results demonstrated that practitioners agreed with 65--72\% of LLMs-generated responses in the IDR case, 83--88\% in the WRS case, 79--86\% in the PHAV case, and 63--76\% in the HRI case. 
In addition, insights from discussions with practitioners suggested that LLMs have the potential to comprehend robots' software requirements, provide logical responses to uncertainties, identify overlooked uncertainties, and demonstrate their usefulness in practical contexts.

This work evaluates LLMs' potential to support practitioners in identifying uncertainties in self-adaptive robotics. We focus on system-level aspects rather than exploring inter-module interactions or specific uncertainty categories. 
Within this scope, the main contributions are: (i) a novel LLM-prompting method for uncertainty identification based on robotic requirements, (ii) an LLM evaluation study with practitioners from four different robotic industries, and (iii) actionable insights and future directions for researchers and practitioners.

\section{Industrial Context}
RoboSAPIENS~\cite{larsen2024robotic} is a European Research and Innovation project developing methods and tools for self-adaptive robotic software, enabling robots to handle unpredictability while ensuring dependability. It connects research and industry partners to create and validate self-adaptive software in real-world use cases. 
RoboSAPIENS implements the MAPLE-K loop (see the left side of \Cref{fig:overview}), an extension of MAPE-K introducing a \textit{Legitimate} phase to ensure planned adaptations comply with safety requirements. Uncertainty is everywhere in a robot's operations, including its environment, hardware issues, and the software implementing the MAPLE-K loop with all its phases. Thus, a systematic method is needed to identify and understand uncertainties when developing self-adaptive robotic software. Below, we briefly describe the four use cases studied in this work.

\subsection{Industrial Disassembly Robot}
This robot focuses on disassembly tasks, specifically dismantling laptops to support refurbishment. Its first task is to unsnap the fitting of a laptop, considering variations and potential challenges of different types of laptops. Next, it unscrews the small bolts one by one, considering screw types, tightness, and destroyed screws. Finally, it removes the screen cable, where tiny pieces of tape must be removed gradually. 

\subsection{Warehouse Robotic Swarm}
This case study involves three omnidirectional mobile robots operating within an indoor delivery system. 
The primary focus is to ensure safe autonomous navigation through individual-level obstacle detection and fleet-level dynamic path planning while considering human presence. 

\subsection{Prolonged Hull of an Autonomous Vessel}
This case focuses on predicting the motion of autonomous vessels during their operation in real and uncertain environments to facilitate dynamic self-adaptation and better decision-making for humans in the loop. 

\subsection{Human-Robotic Interaction}
This case focuses on automating manufacturing by introducing robots as on-demand workers. The robot’s task is to load workpieces into the machine and remove them once processing is complete.

\begin{figure*}
\centerline{\includegraphics[width=\linewidth]{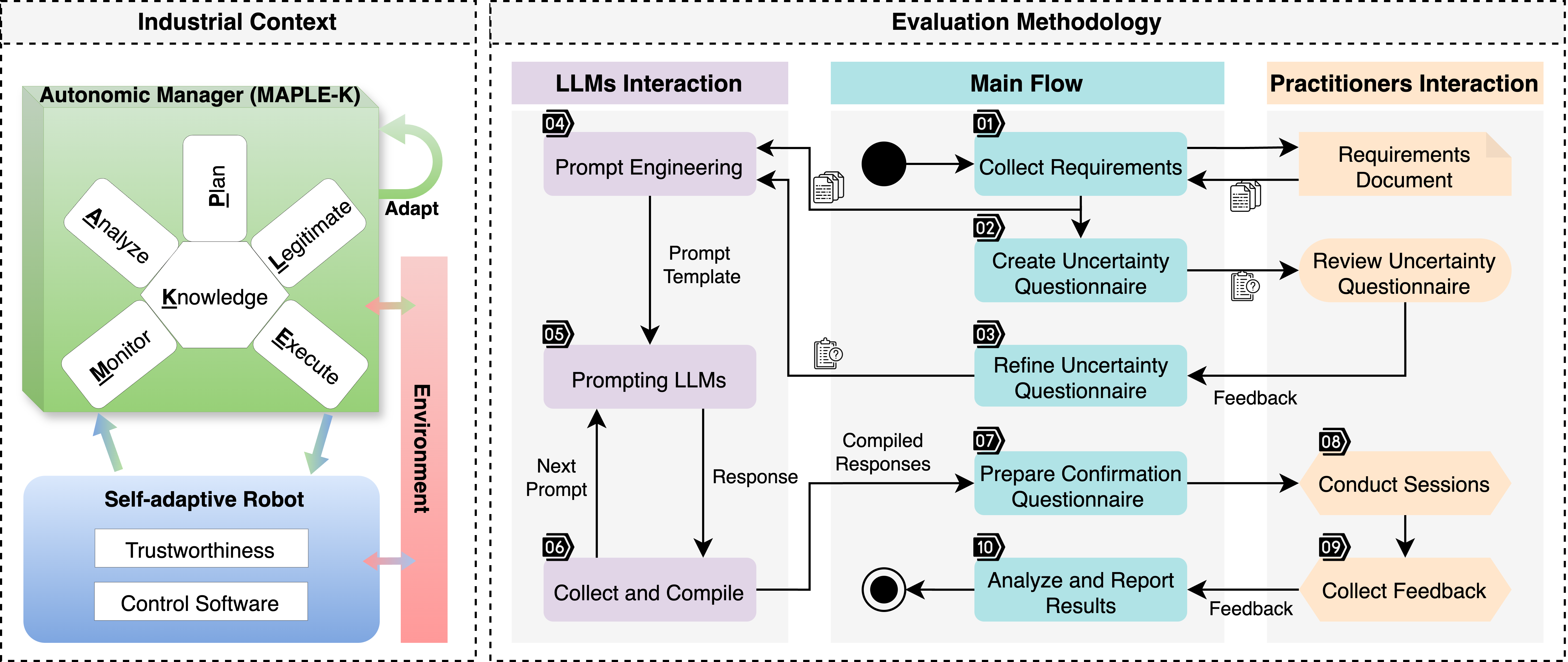}}
\caption{Industrial context (left) and evaluation methodology overview (right). In the workflow, rounded rectangles indicate our activities, ovals represent individual activities by the practitioners, and hexagons denote collaborative activities.}\vspace*{-5pt}
\label{fig:overview}
\end{figure*}

\section{Evaluation Methodology}
\begin{tcolorbox}[colback=red!5!white,colframe=white!75!white,left=3.5pt,right=3.5pt,top=3.5pt,bottom=3.5pt]
\textbf{Goal:} Evaluate LLMs' effectiveness in identifying uncertainties in self-adaptive robotics and to determine their value for practitioners. 
\end{tcolorbox}

\noindent\Cref{fig:overview} presents an overview of our evaluation methodology, encompassing the primary workflow and interactions involving the practitioners and LLMs. 
Below, we elaborate on each step of the process.

\noindent \textbf{Step~\textcolor{blue}{\ding{182}}:}
We collected documented robotic software requirements from the practitioners.  
These documents included detailed descriptions of each use case scenario, clearly specifying the operational context, expected behavior, objectives, and constraints. 
Moreover, they explicitly outlined hardware characteristics, including details of sensors and actuators, as well as thorough software requirements defining the system architecture, functionalities, and performance expectations.

\noindent \textbf{Steps~\textcolor{blue}{\ding{183}}--\textcolor{blue}{\ding{184}}:}
In the second step, we designed an uncertainty questionnaire inspired by existing work~\cite{ramirez2012taxonomy}. 
This questionnaire consisted of seven questions covering various aspects, including uncertainty sources, methods to identify uncertainties, impacted engineering lifecycle phases, the effects of uncertainties on performance and safety, potential mitigation strategies, real-world scenarios where uncertainties resulted in failures, and the overall impact of uncertainties on project outcomes. 
We sent out this questionnaire to the practitioners from all four use cases. 
In the third step, based on the practitioners’ feedback, we revised the questionnaire by rewording certain questions to enhance clarity and comprehensibility. 

\noindent \textbf{Steps~\textcolor{blue}{\ding{185}}--\textcolor{blue}{\ding{187}}:}
The subsequent steps, from four to six, involve interaction with LLMs. 
In step four, we performed prompt engineering to prepare prompts for LLMs. 
We initially tried zero-shot prompting with direct instructions, but the responses lacked precision. 
As our focus is on supporting experts in identifying uncertainties using LLMs, we adopted role-based prompting after exploring zero-shot and self-consistency prompting. 
Role-based prompting lets LLMs assume a specific role (e.g., engineer or tester) to give task-specific responses. 
Therefore, we developed a role-based prompt (available on Zenodo~\cite{LLM4RoboUI}) with step-by-step instructions and the uncertainty questionnaire. 
The instructions include reading requirements, applying robotic knowledge, understanding the questionnaire's purpose, and providing concise, justified responses. 
The next steps involve iteratively prompting all LLMs for each case study and compiling their responses. 
The prompting goal is to obtain answers to uncertainty questions based on their understanding of the robotic context. 
For this, we provided each LLM with the requirement document and the prompt, then collected their responses. 
To mitigate hallucination risks, we used self-consistency prompting by re-prompting the LLMs with the same prompt three times (and up to five times in cases of variability) and selecting the most consistent response. 
After each prompting iteration, final responses were stored and labeled with the corresponding LLM.

\noindent \textbf{Step~\textcolor{blue}{\ding{188}}:}
In the seventh step, to develop confirmation questionnaires, we analyzed all responses to identify unique responses, as we observed that some LLMs provided similar answers to specific questions during prompting. 
We then incorporated all unique responses into the confirmation questionnaire, removing any indication of which LLM generated a particular response. 
The resulting confirmation questionnaire contained all questions from the uncertainty questionnaire and responses from different LLMs.  
For each LLM-generated response, we used a five-point Likert scale to capture practitioners' level of agreement, ranging from ``strongly disagree'' and ``disagree'' to ``neutral'', ``agree'' and ``strongly agree''. 
The questionnaire was developed using Google Forms to facilitate efficient data collection and analysis.
It is important to note that the confirmation questionnaire strictly focused on evaluating the responses provided by LLMs and did not include any questions related to personal information or participants' identities.

\noindent \textbf{Steps~\textcolor{blue}{\ding{189}}--\textcolor{blue}{\ding{190}}:}
In the eighth step, we conducted four individual in-person sessions during project meetings with the practitioners involved in each case study. 
Each session included one senior and one junior participant from the respective case study. 
Each session lasted 40 minutes and followed a consistent format. 
At the beginning of each session, we provided participants with a concise overview of the planned activities and time allocation, presented the revised uncertainty questionnaire, and explained the objectives and motivations of the activity.
Following the 10-minute introductory session, we distributed the confirmation questionnaire to the participants via Google Forms. 
We allocated 20 minutes for the questionnaire and 10 minutes for verbal discussion. 
During the questionnaire activity, participants were asked to assess each response to the uncertainty questionnaire and indicate their level of agreement based on their experience and knowledge of the case study. 
The confirmation questionnaire included sections for each uncertainty question, showing LLM responses and a Likert scale for agreement. 
Participants had sufficient time to review and assess responses thoughtfully. 
They submitted their assessments using Google Forms. 
This was followed by a verbal discussion session to gather broader feedback without revisiting the submitted responses. 
During the verbal discussion, we gathered participants' opinions on the LLM-generated responses, the LLMs' understanding of the robotic case study, and their potential for identifying uncertainties. 
For this discussion, we opted to take informal notes instead of formally recording the feedback.

\noindent \textbf{Step~\textcolor{blue}{\ding{191}}:}
We analyzed the data collected through Google Forms, focusing mainly on feedback from the confirmation questionnaire. 
Our results are primarily derived from the responses to this questionnaire, supplemented by key observations gathered during verbal discussions. 
We analyze the results for each case study, specifically examining the agreement levels expressed by practitioners for each LLM-generated response. 
From this analysis, we identify and report which LLM-generated responses were confirmed by the practitioners, highlighting the levels of consensus and reliability for each response across all use cases.
It is important to note that we obtained informed consent from the participants involved in all case studies to report the evaluation results.

\begin{figure*}
	\subfigure[IDR Results]{\includegraphics[width=8.0cm,height=4.4cm]{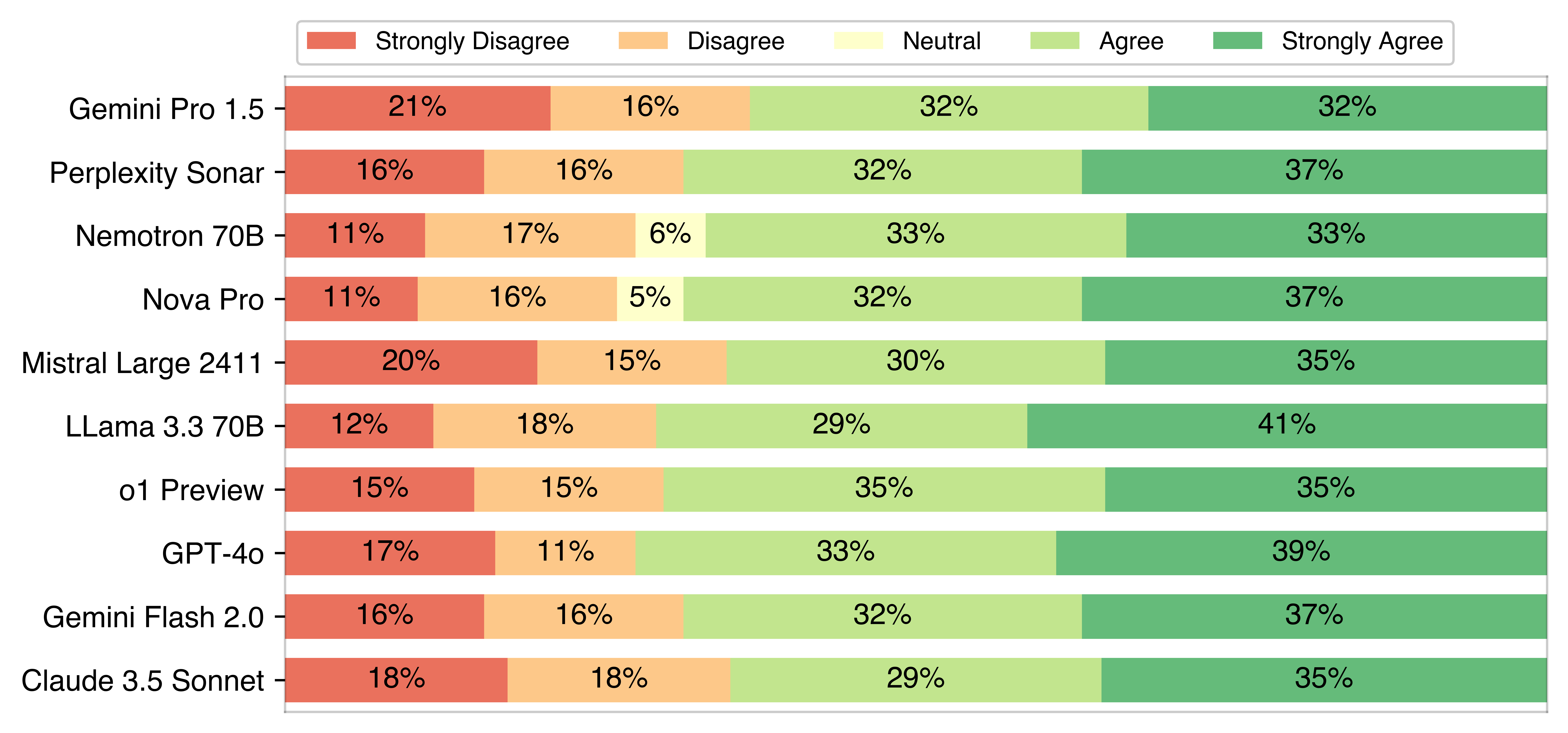}\label{fig:dti}}
	\subfigure[WRS Results]{\includegraphics[width=8.0cm,height=4.4cm]{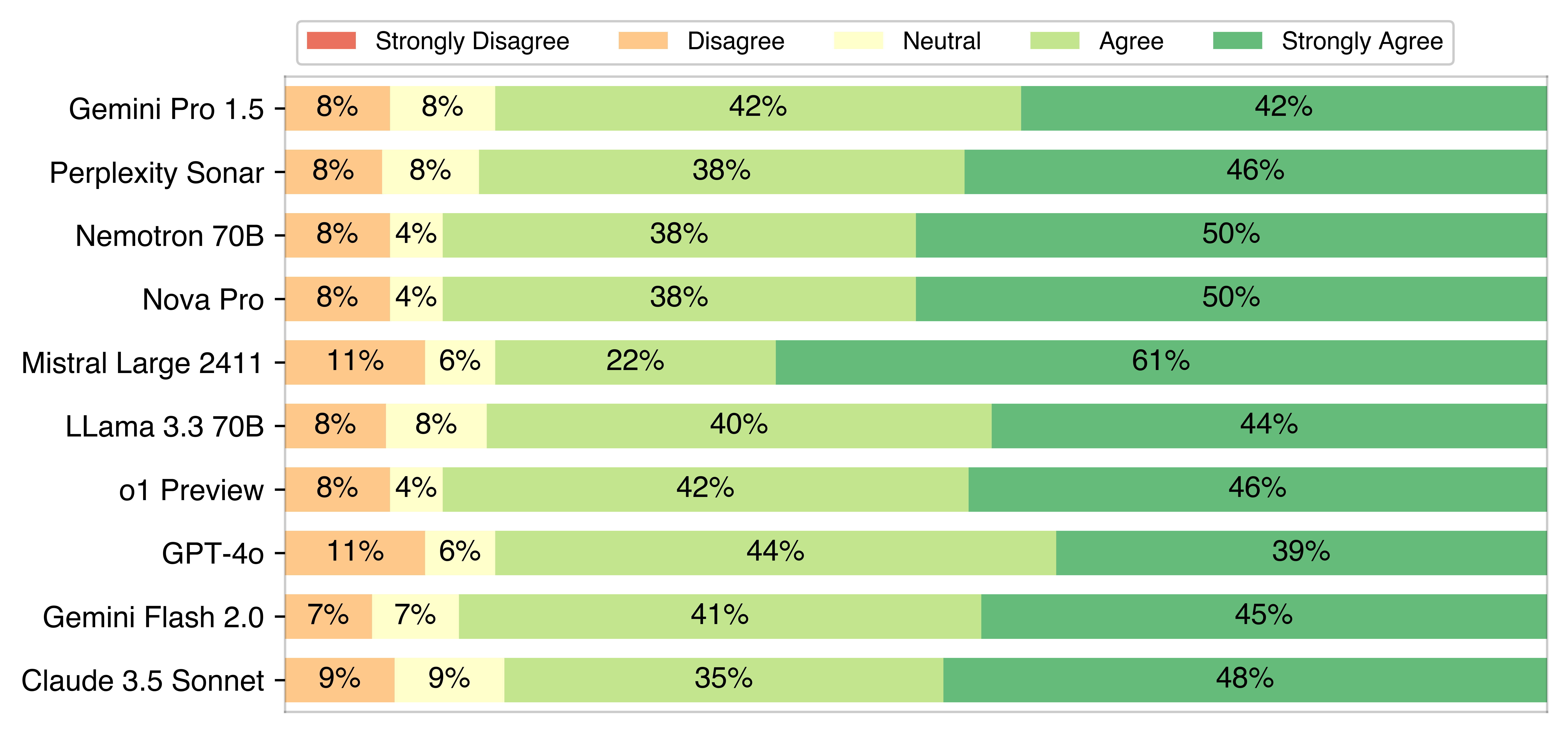}\label{fig:pal}}
	\subfigure[PHAV Results]{\includegraphics[width=8.0cm,height=4.4cm]{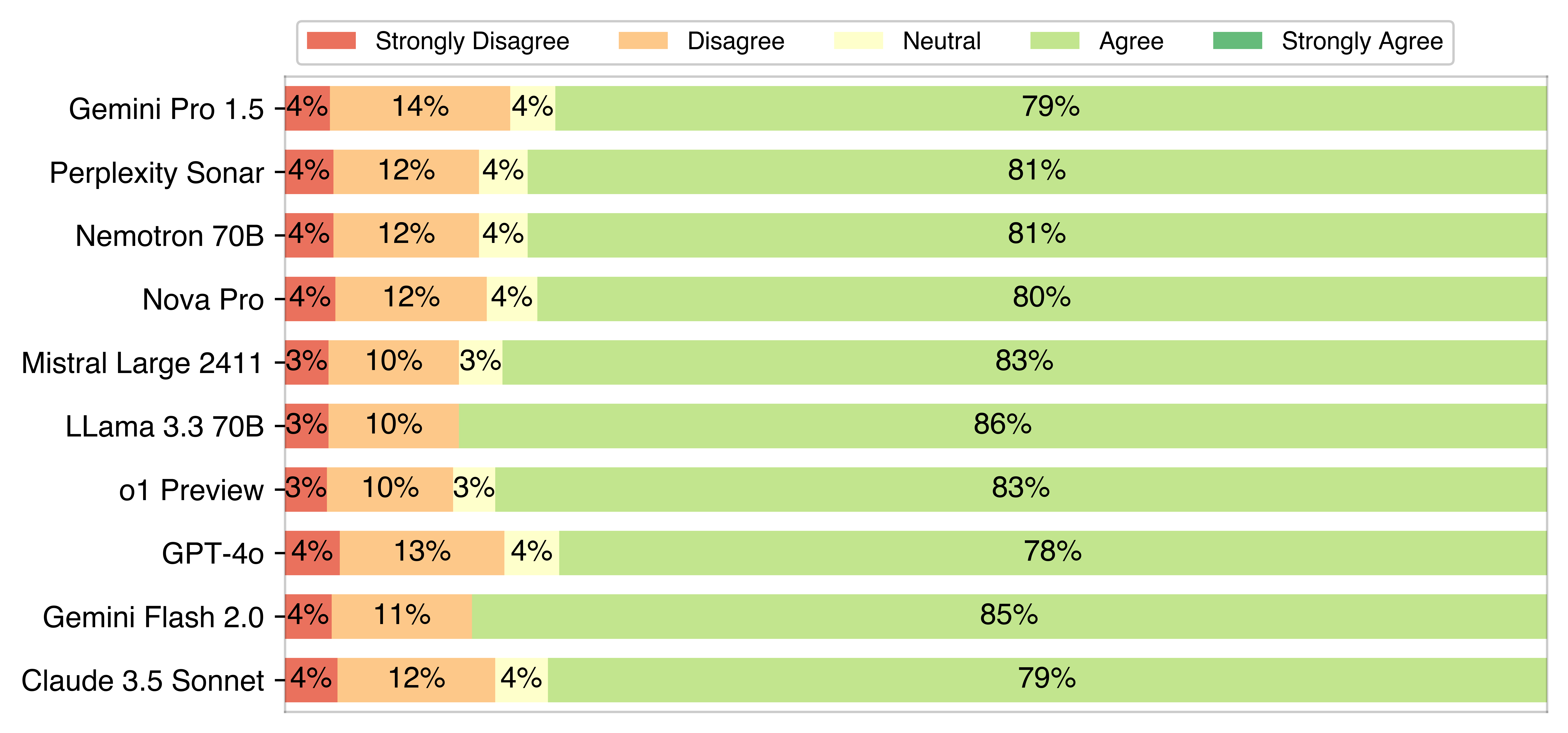}\label{fig:ntnu}}
	\subfigure[HRI Results]{\includegraphics[width=8.0cm,height=4.4cm]{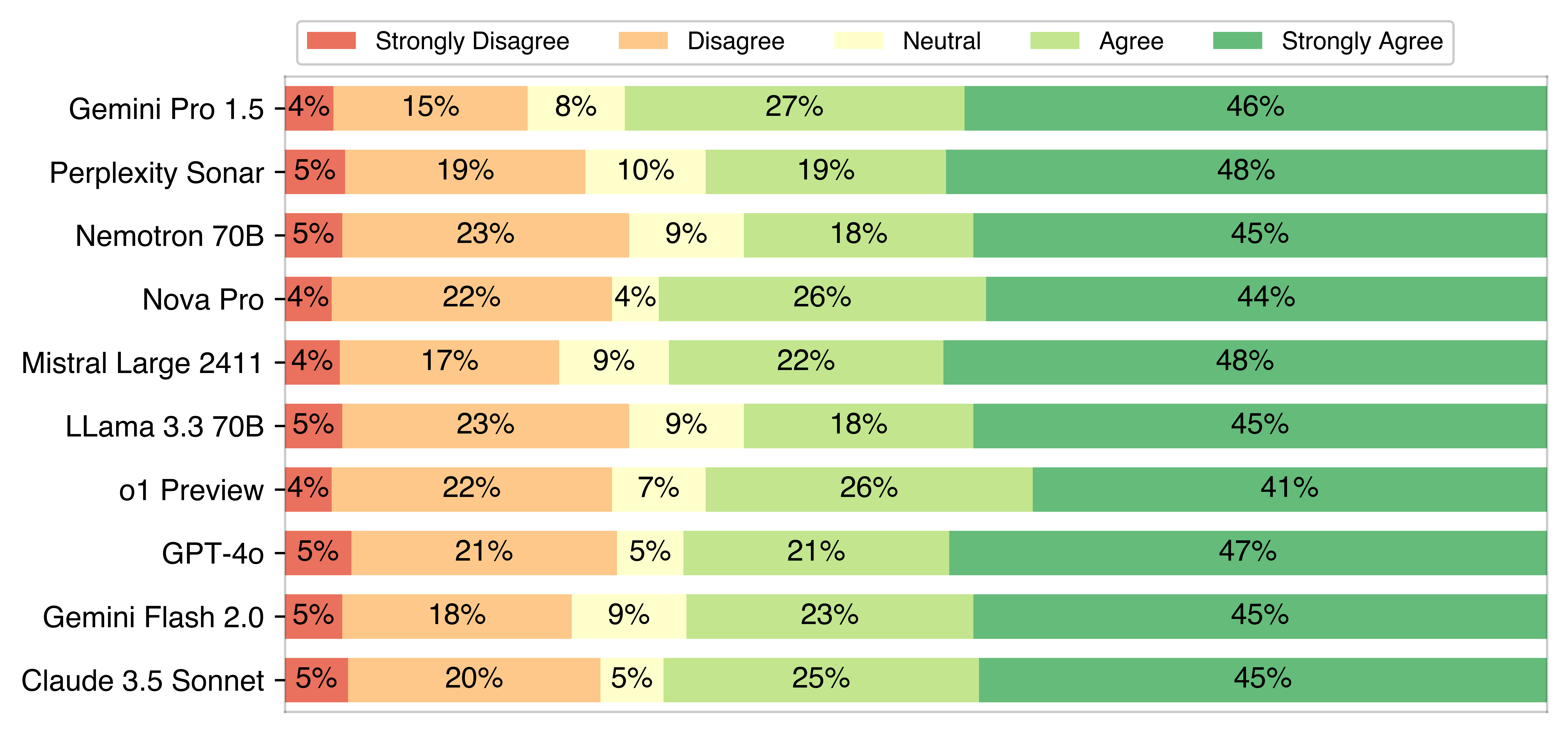}\label{fig:iff}}
    \caption{Results for all LLMs across four case studies, presenting the practitioners’ assessments of LLMs’ effectiveness in identifying uncertainties.} 
    \label{fig:allresults}
\end{figure*}

\section{Results and Insights}
\subsection{IDR Results} 
\Cref{fig:dti} shows that GPT-4o's responses received the highest number of agreements and the lowest disagreements, followed by o1 Preview and LLama 3.3 70B. 
Neutral agreements were reported for only Nemotron 70B (6\%) and Nova Pro (5\%). 
Gemini Pro 1.5, Mistral Large 2411, and Claude 3.5 Sonnet had the highest levels of disagreement. 
Overall, practitioners agreed with over 60\% of the LLM responses, mostly with strong agreements.

\subsection{WRS Results} 
\Cref{fig:pal} demonstrates that practitioners strongly agreed most often with responses generated by Claude 3.5 Sonnet, while overall agreements were high for Nemotron 70B, Nova Pro, and o1 Preview. 
No strong disagreements were reported, with overall disagreements low at 11\%. 
Moreover, neutral agreements are 8\% for responses from Gemini Pro 1.5, Perplexity Sonar, and LLama 3.3 70B. 
In general, practitioners agreed with over 80\% of the LLM responses, highlighting LLMs' high practicality in this case.

\subsection{PHAV Results} 
\Cref{fig:ntnu} indicates that the responses generated by LLama 3.3 70B had the highest agreement and lowest disagreements, followed by Gemini Flash 2.0. 
No neutral agreements were reported for these two models, while Mistral Large 2411 and o1 Preview had 4\% and 3\%, respectively. 
Furthermore, strong disagreements were low overall, while highest for Gemini Pro 1.5. 
Overall, practitioners expressed positive opinions for over 80\% of the LLM responses, highlighting LLMs' high effectiveness.

\subsection{HRI Results} 
\Cref{fig:iff} shows that practitioners strongly agreed with over 40\% of responses from all LLMs, with Perplexity Sonar, Mistral Large 2411, and GPT-4o reaching 48\%, while Nova Pro and o1 Preview received the highest overall agreement. 
For neutral agreements, Perplexity Sonar had the most and Nova Pro the least. 
Moreover, practitioners strongly disagreed with only 4–5\% of LLM responses, while simple disagreements ranged from 15–23\%, with Nemotron 70B and LLama 3.3 70B having the most disagreements and Gemini Pro 1.5 the lowest. 
In general, practitioners agreed with over 70\% of LLM responses, demonstrating LLMs' usefulness in this case.

\subsection{Reliability and Confidence Analysis}
First, we analyzed the inter-rater reliability using Cohen's Kappa, which revealed that both participants unanimously agreed or disagreed in each case, resulting in no false positives or false negatives (detailed results are provided in \Cref{tab:idr,tab:wrs,tab:phav,tab:hri} and discussed in \Cref{app:cohen}). 
Next, we analyzed 95\% confidence intervals using the bootstrap method, which demonstrated high confidence in the results in all case studies (detailed results are presented in \Cref{tab:dti_results,tab:pal_results,tab:ntnu_results,tab:iff_results} and further discussed in \Cref{app:bootstrap}). 
Overall, the strong agreement levels, combined with minimal variation in disagreement and neutrality, demonstrate that the results for each case study are reliable and consistent. 
Detailed findings are also available in the online repository~\cite{LLM4RoboUI}.

\subsection{Insights} 
We present key insights from results analysis and discussions with practitioners.

\subsubsection{LLMs' Familiarity with Domain} 
When we asked practitioners from all case studies whether they believed LLMs could effectively understand the case study specifications and domain, all participants agreed that LLMs' responses indicated a strong grasp of the case study. 
This highlights LLMs’ ability to comprehend text using the provided information about various robots and their prior knowledge from related domains. 

\subsubsection{Logical Responses} 
We asked practitioners about their perspectives on whether LLMs’ responses were logical and contextually relevant. 
All participants agreed that the responses were logical and aligned with the given context. 
This demonstrates LLMs' analytical capability and their usefulness in identifying uncertainties in self-adaptive robotics. 

\subsubsection{Overlooked Uncertainties} 
We asked practitioners whether they found the uncertainties identified by LLMs to be realistic and whether any of them were previously unknown yet logically valid. 
They unanimously agreed that the LLM-identified uncertainties were realistic and were particularly surprised by some previously unconsidered uncertainties, which they found valuable. 
Below, we present two uncertainty examples for each case study, classified by type, source, lifecycle stage, and alignment with existing taxonomies. 
In IDR, variations in laptop components can be screws (aleatoric; hardware; operational; corresponds~\cite{ramirez2012taxonomy}) and cable routing issues during disassembly (epistemic; internal; operational; corresponds~\cite{Hezavehi21}). 
In WRS, LiDAR sensor occlusion may disrupt robot navigation (aleatoric; environmental; operational; corresponds~\cite{ramirez2012taxonomy}) and mislocalize due to odometry errors (epistemic; software/internal; operational; corresponds~\cite{Hezavehi21}). 
In PHAV, uncertainty can result from transferring models between vessels with differing dynamics (epistemic; software/internal; testing; corresponds~\cite{Hezavehi21}) and unexpected weather changes with complex wave-current interactions (aleatoric; environmental; operational; corresponds~\cite{ramirez2012taxonomy}).  
In HRI, unnecessary interactions with humans and objects may cause frequent re-planning (aleatoric; external; operational; corresponds~\cite{Harriet2024}) and variations in the workspace setup or additional obstacles (epistemic; environmental/external; operational; corresponds~\cite{Hezavehi21}).

\subsubsection{Knowledge Gaps in Engineering Phases} 
Practitioners revealed a lack of knowledge about uncertainties across different robotic software engineering phases. 
Some participants focused only on development, others on testing, or operations. 
Developers had limited awareness of operational uncertainties, while those in the operations phase were unfamiliar with design-phase uncertainties. 
Consequently, they could not confirm uncertainties outside their expertise and either disagreed or remained neutral when evaluating LLM-identified uncertainties related to phases they were not directly involved. 

\subsubsection{Uncertainty in Engineering Phases:} 
Although all practitioners agreed that uncertainties arise in every engineering phase, the majority emphasized that the most uncertainties are typically encountered during the testing phase. 
Those involved in the operational phase highlighted that uncertainties are equally prevalent during live operations. 
This was especially noticeable in the autonomous vessel case study, where real-world operations in complex environments often introduce significant environmental uncertainties that may go undetected during simulation-based testing. 
Therefore, depending on the complexity and dynamics of the environment, both the testing and operational phases are likely to encounter uncertainties. 
Furthermore, all practitioners unanimously confirmed that environmental dynamics are a major source of uncertainty. 

\subsubsection{Managing Uncertainty} 
Our analysis revealed that the techniques mostly used to manage uncertainty are modeling, simulation, digital twins, and uncertainty quantification. 
However, we also noticed that applying these techniques, monitoring uncertain scenarios, and taking countermeasures require significant manual effort.  
This necessitates developing frameworks and tools to support automation. 
In this context, we foresee significant potential for LLMs, given the usefulness demonstrated in our evaluation.

\subsubsection{LLM Uncertainty Identification Trends}
LLMs effectively identify uncertainties from detailed hardware and software specifications, as evident from the WRS results. 
However, LLMs often apply their general uncertainty knowledge to all aspects mentioned in the requirements, which may not fit specific systems. 
For example, uncertainties from human error are less relevant for autonomous vessels (PHAVs) due to experienced operators and well-defined protocols, though such uncertainties may arise in WRS. 
Regarding uncertainty occurrence, uncertainties arise primarily from hardware issues and dynamic environments, followed by ML models in MAPLE-K.

\subsubsection{LLM Reasoning}
While we observed no major differences among LLMs with varying reasoning capabilities, reasoning models like o1 Preview generated more detailed and well-structured responses, whereas non-reasoning models like Perplexity Sonar still attempted to incorporate some level of reasoning in their outputs. 
Furthermore, general-purpose models like GPT-4o often generated broader, less context-specific reasoning, which frequently led to practitioner disagreement. 
In particular, the example uncertainty scenarios generated by the reasoning models were interesting and endorsed by practitioners. 
Therefore, we recommend using reasoning models with advanced prompting techniques for effective uncertainty analysis.

\subsubsection{Actionable Recommendations}
Given LLMs' varied reasoning capabilities, practitioners shall adopt a combination of prompting strategies, such as few-shot prompting for providing examples and chain-of-thought prompting for step-by-step reasoning, to iteratively refine responses and enhance the identification of true uncertainties. 
To minimize hallucination risks, practitioners identifying uncertainties at the design phase should communicate LLM responses with teams across engineering phases, such as operations and testing, and incorporate their feedback into subsequent prompts. 
As best practices, practitioners should use a second LLM to cross-check responses for alternative perspectives and greater confidence in the results, and remove sensitive information from documents provided to LLMs.

\subsection{Threats to Validity} 
A potential threat to external validity is the limited sample size of eight practitioners involved in four robotic cases focused on self-adaptive behaviors. 
However, involving more participants was infeasible due to a limited number of personnel allocated to the project, so we included those with sufficient experience and familiarity with robotics to address this. 
Although the results may not generalize to all robotic systems, such limitations are common in empirical studies involving industry collaborations. 
An internal validity threat may occur due to participants' personal biases and familiarity with robotic systems. 
Although our sessions with practitioners involved two participants per case (one junior and one senior), their backgrounds and confidence levels may have influenced their responses. 
To handle this, we included open discussions during sessions to clarify interpretations. 
A potential conclusion validity threat may occur due to variability in practitioners' experience levels and the subjective nature of agreement measurements (i.e., Likert levels). 
To mitigate this, we used standardized procedures in all sessions, providing clear instructions and consistent timing, although subjective assessment remains an inherent challenge. 
Moreover, we used Cohen's Kappa to analyze inter-rater reliability and bootstrap confidence intervals to evaluate the confidence in the results. 
A possible threat involves the clarity of the uncertainty questionnaire and confirmation questionnaire. 
To manage this, we developed the uncertainty questionnaire based on existing frameworks and revised it based on practitioners' feedback. 
We also used the standard Likert scale to capture agreement levels for the confirmation questionnaire precisely.

\section{Related Works} 
Despite tremendous research efforts on identifying uncertainty in self-adaptive robots \cite{zheng2024,Hezavehi21,Harriet2024,SUK2024317,Chirayil24}, a systematic approach to identifying and mitigating uncertainties is still lacking \cite{Busch2022}. These studies focus on specific types, sources, and dimensions of uncertainty separately, overlooking complex interdependencies and environmental correlations~\cite{Wang2025}. 
Hezavehi \textit{et al.}~\cite{Hezavehi21} conducted a field survey and identified that uncertainties in self-adaptive systems are mostly application-dependent and rely on non-functional requirements. 
Chirayil \textit{et al.}~\cite{Chirayil24} proposed a unified classification of anomalies for autonomous robotic missions. 
Similarly, Busch \textit{et al.}~\cite{Busch2022} proposed a setup for identifying and quantifying the uncertainty of the eigenfrequency in machining robots. 
Recently, Zheng \textit{et al.} \cite{zheng2024} developed an uncertainty-based LLM failure detector for autonomous robots to enable efficient task planning. 
Moreover, Betzer \textit{et al.}~\cite{Betzer2024} developed a cloud-based digital twin to enable real-time identification and mitigation of uncertainties for an autonomous mobile robot. 
Compared to these studies, our work explores LLMs' potential in systematically identifying uncertainties in self-adaptive robotics.

\section{Conclusion}
We evaluated the LLMs' potential to support uncertainty identification in four industrial-sized self-adaptive robotic cases using 10 LLMs.  
We provided these LLMs with case study requirements and a prompt that contained an uncertainty questionnaire. 
The responses generated by the LLMs were then evaluated with input from the practitioners. 
Results demonstrated that practitioners agreed with 63--88\% of the LLM responses. 
Furthermore, discussions with practitioners highlighted the usefulness of LLMs in understanding robotic cases, generating logical responses, and identifying overlooked uncertainties.

Our study is a first step in demonstrating LLMs' potential to identify uncertainties and opens several future research directions. 
One direction is to conduct a study with domain experts and perform a comparative analysis of expert-identified uncertainties with those generated by LLMs. 
Another is to conduct a large-scale empirical evaluation involving multiple robotic systems and participants with different levels of expertise, with follow-up experiments or uncertainty simulations to validate the findings. 
Furthermore, future work could focus on developing a holistic uncertainty identification framework and constructing an uncertainty taxonomy inspired by empirical findings and existing literature.

\section*{Acknowledgments}
This work is funded by the RoboSAPIENS project under the EU Horizon Europe program (Grant No. 101133807).

\bibliographystyle{unsrt}  
\bibliography{refs}  

\begin{thebibliography}{10}

\bibitem{larsen2024robotic}
Peter~G Larsen, Shaukat Ali, Roland Behrens, Ana Cavalcanti, Claudio Gomes, Guoyuan Li, Paul De~Meulenaere, Mikkel~L Olsen, Nikolaos Passalis, Thomas Peyrucain, et~al.
\newblock {Robotic safe adaptation in unprecedented situations: the RoboSAPIENS project}.
\newblock {\em Research Directions: Cyber-Physical Systems}, 2:e4, 2024.

\bibitem{kephart2003vision}
Jeffrey~O Kephart and David~M Chess.
\newblock The vision of autonomic computing.
\newblock {\em Computer}, 36(1):41--50, 2003.

\bibitem{weyns2023towards}
Danny Weyns, Radu Calinescu, Raffaela Mirandola, Kenji Tei, Maribel Acosta, Nelly Bencomo, Amel Bennaceur, Nicolas Boltz, Tomas Bures, Javier Camara, et~al.
\newblock Towards a research agenda for understanding and managing uncertainty in self-adaptive systems.
\newblock {\em SIGSOFT Software Engineering Notes}, 48(4):20--36, 2023.

\bibitem{LMMsDawn2023}
Zhengyuan Yang, Linjie Li, Kevin Lin, Jianfeng Wang, Chung-Ching Lin, Zicheng Liu, and Lijuan Wang.
\newblock {The Dawn of LMMs: Preliminary Explorations with GPT-4V(ision)}, 2023.

\bibitem{hassan2024rethinking}
Ahmed~E. Hassan, Gustavo~A. Oliva, Dayi Lin, Boyuan Chen, Zhen Ming, and Jiang.
\newblock {Rethinking Software Engineering in the Foundation Model Era: From Task-Driven AI Copilots to Goal-Driven AI Pair Programmers}, 2024.

\bibitem{ramirez2012taxonomy}
Andres~J Ramirez, Adam~C Jensen, and Betty~HC Cheng.
\newblock A taxonomy of uncertainty for dynamically adaptive systems.
\newblock In {\em 2012 7th International Symposium on Software Engineering for Adaptive and Self-Managing Systems (SEAMS)}, pages 99--108. IEEE, 2012.

\bibitem{LLM4RoboUI}
Hassan Sartaj.
\newblock {LLM4RoboUI}: {LLM}-based uncertainty identification in self-adaptive robotics, October 2025.

\bibitem{Hezavehi21}
Sara~M. Hezavehi, Danny Weyns, Paris Avgeriou, Radu Calinescu, Raffaela Mirandola, and Diego Perez-Palacin.
\newblock Uncertainty in self-adaptive systems: A research community perspective.
\newblock {\em ACM Transactions on Autonomous and Adaptive Systems (TAAS)}, 15(4):1--36, 2021.

\bibitem{Harriet2024}
Harriet~R. Cameron, Simon Castle-Green, Muhammad Chughtai, Liz Dowthwaite, Ayse Kucukyilmaz, Horia~A. Maior, Victor Ngo, Eike Schneiders, and Bernd~C. Stahl.
\newblock A taxonomy of domestic robot failure outcomes: Understanding the impact of failure on trustworthiness of domestic robots.
\newblock In {\em Proceedings of the Second International Symposium on Trustworthy Autonomous Systems}, TAS '24, pages 1--14, New York, NY, USA, 2024. Association for Computing Machinery.

\bibitem{zheng2024}
Zhi Zheng, Qian Feng, Hang Li, Alois Knoll, and Jianxiang Feng.
\newblock Evaluating uncertainty-based failure detection for closed-loop llm planners, 2025.

\bibitem{SUK2024317}
Ho~Suk, Yerin Lee, Taewoo Kim, and Shiho Kim.
\newblock Addressing uncertainty challenges for autonomous driving in real-world environments.
\newblock In {\em Advances in computers}, volume 134, pages 317--361. Elsevier, 2024.

\bibitem{Chirayil24}
Shivoh Chirayil~Nandakumar, Daniel Mitchell, Mustafa~Suphi Erden, David Flynn, and Theodore Lim.
\newblock Anomaly detection methods in autonomous robotic missions.
\newblock {\em Sensors}, 24(4):1330, 2024.

\bibitem{Busch2022}
Maximilian Busch, Florian Schnoes, Amr Elsharkawy, and Michael~F Zaeh.
\newblock Methodology for model-based uncertainty quantification of the vibrational properties of machining robots.
\newblock {\em Robotics and Computer-Integrated Manufacturing}, 73:102243, 2022.

\bibitem{Wang2025}
Ke~Wang, Chongqiang Shen, Xingcan Li, and Jianbo Lu.
\newblock Uncertainty quantification for safe and reliable autonomous vehicles: A review of methods and applications.
\newblock {\em IEEE Transactions on Intelligent Transportation Systems}, 26(3):2880--2896, 2025.

\bibitem{Betzer2024}
Joakim~Schack Betzer, Jalil Boudjadar, Mirgita Frasheri, and Prasad Talasila.
\newblock Digital twin enabled runtime verification for autonomous mobile robots under uncertainty.
\newblock In {\em International Symposium on Distributed Simulation and Real Time Applications}, pages 10--17. IEEE, 2024.

\end{thebibliography}

\appendix
\section*{Appendix: Supplementary Findings}

\section{Reliability Analysis of Results using Cohen’s K}\label{app:cohen}
In the following, we present the results of the reliability analysis for all four case studies.

\subsection*{Case 1: Industrial Disassembly Robot (IDR)}
\Cref{tab:idr} presents inter-rater reliability results obtained using Cohen’s K measure. 
The results reveal unanimous agreement between participants on both \textit{agree} and \textit{disagree} decisions.
Therefore, no false positives or false negatives were observed.

\begin{table}[htb]
\centering
\caption{IDR case study results for inter-rater agreement analysis with Cohen’s K}
\label{tab:idr}
\begin{tabular}{l|lll|}
\cline{2-4}
 & \multicolumn{3}{c|}{\textbf{Rater 2}} \\ \hline
\multicolumn{1}{|l|}{\multirow{3}{*}{\textbf{Rater 1}}} & \multicolumn{1}{l|}{} & \multicolumn{1}{l|}{Agree} & Disagree \\ \cline{2-4} 
\multicolumn{1}{|l|}{} & \multicolumn{1}{l|}{Agree} & \multicolumn{1}{l|}{126} & 0 \\ \cline{2-4} 
\multicolumn{1}{|l|}{} & \multicolumn{1}{l|}{Disagree} & \multicolumn{1}{l|}{0} & 58 \\ \hline
\end{tabular}
\end{table}

\subsection*{Case 2: Warehouse Robotic Swarm (WRS)}
\Cref{tab:wrs} shows inter-rater reliability results calculated using Cohen’s K measure. 
The results indicate a unanimous agreement between both participants on \textit{agree} and \textit{disagree}.
Hence, there are no false positives or false negatives in this case.

\begin{table}[htb]
\centering
\caption{WRS case study results for inter-rater agreement analysis with Cohen’s K}
\label{tab:wrs}
\begin{tabular}{l|lll|}
\cline{2-4}
 & \multicolumn{3}{c|}{\textbf{Rater 2}} \\ \hline
\multicolumn{1}{|l|}{\multirow{3}{*}{\textbf{Rater 1}}} & \multicolumn{1}{l|}{} & \multicolumn{1}{l|}{Agree} & Disagree \\ \cline{2-4} 
\multicolumn{1}{|l|}{} & \multicolumn{1}{l|}{Agree} & \multicolumn{1}{l|}{200} & 0 \\ \cline{2-4} 
\multicolumn{1}{|l|}{} & \multicolumn{1}{l|}{Disagree} & \multicolumn{1}{l|}{0} & 20 \\ \hline
\end{tabular}
\end{table}

\subsection*{Case 3: Prolonged Hull of an Autonomous Vessel (PHAV)}
\Cref{tab:phav} presents inter-rater reliability results obtained using Cohen’s K measure. 
The results indicate unanimous agreement between participants on both \textit{agree} and \textit{disagree} choices.
Thus, no false positives or false negatives were observed.

\begin{table}[htb]
\centering
\caption{PHAV case study results for inter-rater agreement analysis with Cohen’s K}
\label{tab:phav}
\begin{tabular}{l|lll|}
\cline{2-4}
 & \multicolumn{3}{c|}{\textbf{Rater 2}} \\ \hline
\multicolumn{1}{|l|}{\multirow{3}{*}{\textbf{Rater 1}}} & \multicolumn{1}{l|}{} & \multicolumn{1}{l|}{Agree} & Disagree \\ \cline{2-4} 
\multicolumn{1}{|l|}{} & \multicolumn{1}{l|}{Agree} & \multicolumn{1}{l|}{218} & 0 \\ \cline{2-4} 
\multicolumn{1}{|l|}{} & \multicolumn{1}{l|}{Disagree} & \multicolumn{1}{l|}{0} & 41 \\ \hline
\end{tabular}
\end{table}

\subsection*{Case 4: Human-Robotic Interaction (HRI)}
\Cref{tab:hri} shows inter-rater reliability results obtained using Cohen’s K measure. 
The results demonstrate that both participants unanimously \textit{agreed} and \textit{disagreed}.
Consequently, no false positives or false negatives exist.

\begin{table}[htb]
\centering
\caption{HRI case study results for inter-rater agreement analysis with Cohen’s K}
\label{tab:hri}
\begin{tabular}{l|lll|}
\cline{2-4}
 & \multicolumn{3}{c|}{\textbf{Rater 2}} \\ \hline
\multicolumn{1}{|l|}{\multirow{3}{*}{\textbf{Rater 1}}} & \multicolumn{1}{l|}{} & \multicolumn{1}{l|}{Agree} & Disagree \\ \cline{2-4} 
\multicolumn{1}{|l|}{} & \multicolumn{1}{l|}{Agree} & \multicolumn{1}{l|}{160} & 0 \\ \cline{2-4} 
\multicolumn{1}{|l|}{} & \multicolumn{1}{l|}{Disagree} & \multicolumn{1}{l|}{0} & 56 \\ \hline
\end{tabular}
\end{table}

\section{Results for Confidence Intervals with Bootstrap Method}\label{app:bootstrap}
In the following, we present the results of non-parametric confidence intervals calculated using the bootstrap method for all four case studies. 

\subsection*{Case 1: Industrial Disassembly Robot (IDR)}
The results in~\Cref{tab:dti_results} provide 95\% bootstrap confidence intervals (CI) for each agreement category across the evaluated models. The \textbf{Strongly Disagree} category shows a relatively narrow confidence interval (2.40–3.40), indicating consistent responses across models. Similarly, the \textbf{Agree} and \textbf{Strongly Agree} categories show higher and stable agreement levels, with confidence intervals of 5.60–6.40 and 6.00–7.00, respectively. The \textbf{Neutral} category has the smallest confidence interval (0.00–0.40), reflecting minimal neutral responses across the models, while \textbf{Disagree} exhibits a narrow range (2.80–3.00), suggesting low disagreement levels. Overall, the results highlight that most models are rated positively, with strong agreement dominating the responses.

\begin{table}[htb]
\centering
\caption{Confidence Intervals for IDR Results}
\label{tab:dti_results}
\begin{tabular}{|l|c|c|}
\hline
\textbf{Category} & \textbf{Lower Bound (95\% CI)} & \textbf{Upper Bound (95\% CI)} \\
\hline
Strongly Disagree & 2.40 & 3.40 \\
Disagree & 2.80 & 3.00 \\
Neutral & 0.00 & 0.40 \\
Agree & 5.60 & 6.40 \\
Strongly Agree & 6.00 & 7.00 \\
\hline
\end{tabular}
\end{table}

\subsection*{Case 2: Warehouse Robotic Swarm (WRS)}

\Cref{tab:pal_results} presents 95\% bootstrap confidence intervals for the mean responses across five Likert-scale categories in the WRS results. The \textbf{Strongly Disagree} and \textbf{Disagree} categories show narrow confidence intervals ([0.00, 0.00] and [2.00, 2.00], respectively), indicating consistent responses across the models with minimal variation. In contrast, the \textbf{Neutral} category has a slightly broader interval ([1.10, 1.90]), reflecting moderate variability in responses. The \textbf{Agree} and \textbf{Strongly Agree} categories exhibit higher average values with intervals of [7.60, 10.20] and [10.10, 11.90], respectively, highlighting a strong positive agreement among the models. These results suggest that the majority of the responses are skewed toward agreement, with minimal disagreement or neutrality.

\begin{table}[htb]
\centering
\caption{Confidence Intervals for WRS Results}
\label{tab:pal_results}
\begin{tabular}{|l|c|c|}
\hline
\textbf{Category} & \textbf{Lower Bound (95\% CI)} & \textbf{Upper Bound (95\% CI)} \\
\hline
Strongly Disagree & 0.00 & 0.00 \\
Disagree & 2.00 & 2.00 \\
Neutral & 1.10 & 1.90 \\
Agree & 7.60 & 10.20 \\
Strongly Agree & 10.10 & 11.90 \\
\hline
\end{tabular}
\end{table}

\subsection*{Case 3: Prolonged Hull of an Autonomous Vessel (PHAV)}
\Cref{tab:ntnu_results} shows 95\% bootstrap confidence intervals for the five Likert-scale categories in the PHAV results. The \textbf{Strongly Disagree} and \textbf{Strongly Agree} categories have narrow confidence intervals ([1.00, 1.00] and [0.00, 0.00], respectively), indicating highly consistent responses across models with minimal variation. The \textbf{Agree} category exhibits the widest interval ([20.20, 23.20]), reflecting a strong positive agreement but with moderate variability in the responses. Meanwhile, the \textbf{Neutral} category has a relatively small interval ([0.70, 1.10]), suggesting limited neutral responses across the models. Overall, the results indicate that most responses are skewed towards agreement, with little disagreement or neutrality observed.

\begin{table}[htb]
\centering
\caption{Confidence Intervals for PHAV Results}
\label{tab:ntnu_results}
\begin{tabular}{|l|c|c|}
\hline
\textbf{Category} & \textbf{Lower Bound (95\% CI)} & \textbf{Upper Bound (95\% CI)} \\
\hline
Strongly Disagree & 1.00 & 1.00 \\
Disagree & 3.00 & 3.10 \\
Neutral & 0.70 & 1.10 \\
Agree & 20.20 & 23.20 \\
Strongly Agree & 0.00 & 0.00 \\
\hline
\end{tabular}
\end{table}

\subsection*{Case 4: Human-Robotic Interaction (HRI)}
\Cref{tab:iff_results} presents 95\% bootstrap confidence intervals for the five Likert-scale categories in the HRI results. The \textbf{Strongly Disagree} category shows a very narrow confidence interval ([1.00, 1.00]), indicating consistent responses across all models. The \textbf{Disagree} and \textbf{Neutral} categories exhibit slightly broader intervals ([4.30, 5.30] and [1.30, 2.00], respectively), reflecting moderate variability in responses. The \textbf{Agree} category shows a wider interval ([4.40, 6.60]), suggesting variability in agreement levels across models. Finally, the \textbf{Strongly Agree} category has a high interval ([9.60, 11.20]), indicating strong agreement among the models with some variability. Overall, the results highlight consistent disagreement and strong agreement trends across most models.

\begin{table}[htb]
\centering
\caption{Confidence Intervals for HRI Results}
\label{tab:iff_results}
\begin{tabular}{|l|c|c|}
\hline
\textbf{Category} & \textbf{Lower Bound (95\% CI)} & \textbf{Upper Bound (95\% CI)} \\
\hline
Strongly Disagree & 1.00 & 1.00 \\
Disagree & 4.30 & 5.30 \\
Neutral & 1.30 & 2.00 \\
Agree & 4.40 & 6.60 \\
Strongly Agree & 9.60 & 11.20 \\
\hline
\end{tabular}
\end{table}

\end{document}